\documentclass{article}
\usepackage{spconf,amsmath,graphicx}
\usepackage{enumitem}
\setlist{nosep}

\usepackage{subcaption}
\usepackage[hidelinks]{hyperref}
\usepackage{multirow}

\newcommand\blfootnote[1]{%
  \begingroup
  \renewcommand\thefootnote{}\footnote{#1}%
  \addtocounter{footnote}{-1}%
  \endgroup
}

\title{Deep learning classification of large-scale point clouds: \\ A case study on cuneiform tablets}

\name{Frederik Hagelskjær}
\address{SDU Robotics, University of Southern Denmark\\
\texttt{frhag@mmmi.sdu.dk}}

\begin{document}
%


\maketitle

\begin{abstract}
This paper introduces a novel network architecture for the classification of large-scale point clouds. 
The network is used to classify metadata from cuneiform tablets. 
As more than half a million tablets remain unprocessed, this can help create an overview of the tablets.
The network is tested on a comparison dataset and obtains state-of-the-art performance. 
We also introduce new metadata classification tasks on which the network shows promising results.
Finally, we introduce the novel Maximum Attention visualization, demonstrating that the trained network focuses on the intended features.
Code available at \url{https://github.com/fhagelskjaer/dlc-cuneiform}
\end{abstract}

\section{Introduction}
\label{sec:intro}
\blfootnote{This project was funded in part by Innovation Fund Denmark through the project MADE FAST, in part by the SDU I4.0-Lab.}
The oldest recorded writing is cuneiform signs imprinted on clay tablets.
Cuneiform writing was used at least from the fourth millennia BCE until the end of the first millennia BCE \cite{walker1987cuneiform}.
As a result of the durability of the clay, vast quantities of tablets have been excavated during the years.
However, the number of assyriologists to process these tablets does not match the massive amount of excavated tablets.
This is emphasized by the fact that there are currently more than half a million unprocessed tablets \cite{cohen2004iclay}.
As these tablets contain vast amounts of knowledge currently unavailable, it is very desirable to process cuneiform tablets automatically.
Methods have demonstrated that deep learning is possible with cuneiform data, but these methods either require manual preprocessing \cite{zampieri2019report} or only use 2D scans \cite{gordin2020reading}.
%
However, as cuneiform tablets are 3D objects, the text often wraps around the corners \cite[p.~14]{walker1987cuneiform}.
The processing, therefore, needs to be in 3D to include all information \cite{bogacz2020period}.
As a result, datasets have been created of 3D scanned tablets \cite{cohen2004iclay, mara2019breaking, anderson2002unwrapping}.
Deep learning, therefore, needs to be performed in 3D to process the tablets thoroughly.
%
%
%
%
%
The introduction of the PointNet architecture demonstrated that deep learning could be performed on 3D models by conversion to point clouds \cite{qi2017pointnet}.
Since then, a number of architectures for deep learning of point clouds have been introduced \cite{guo2020deep}. While successful, these networks generally handle smaller point clouds \cite{bello2020deep}.
DGCNN handles point clouds of up to 4096 points \cite{dgcnn} and PointNet++ \cite{qi2017pointnetplusplus} go to 8192 points. 
However, to include the fine-grained details of the tablets, large-scale point clouds are needed. 
In our method, we process point clouds of 32,768 points.%
%
%

\begin{figure}[t]
    \begin{center}
        \includegraphics[width=0.33\linewidth]{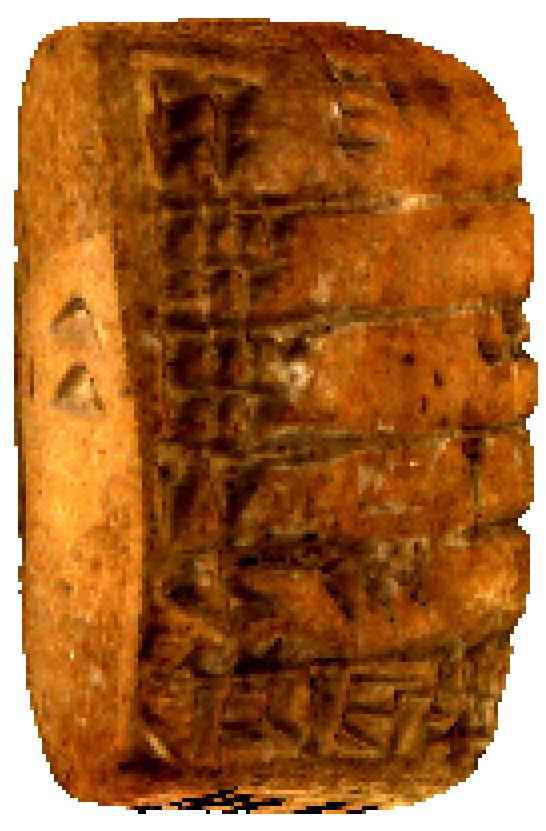}
        ~~~~~~
        ~
        \includegraphics[width=0.33\linewidth]{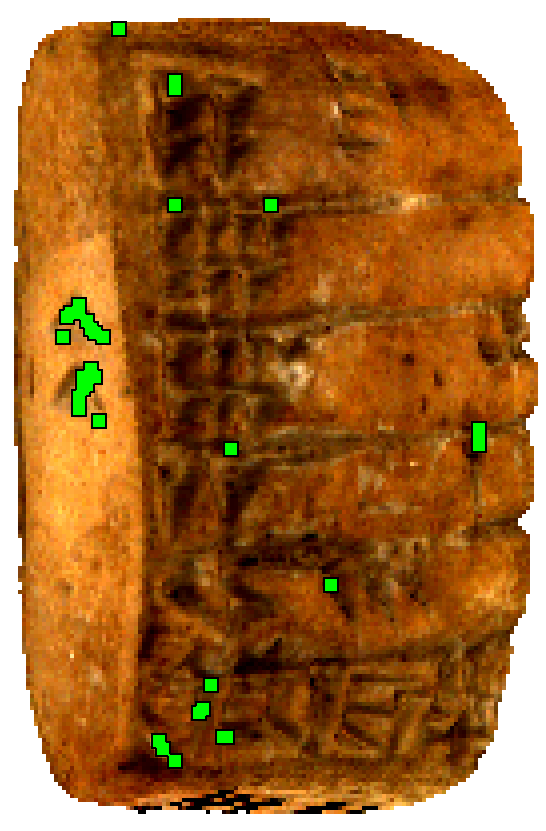}
        \caption{\textbf{Left:} Tablet "HS 2275" \cite{mara2019breaking}. \textbf{Right:} Our Maximum Attention visualization shows the positive attention of the network. The network is trained to classify tablets with signs on the left side, and most of the attention is around this sign.}
        \label{fig:activation}
        \vspace{-5mm}
    \end{center}
\end{figure}
A successful network for processing very large point clouds is introduced by Bogacz et al. \cite{bogacz2020period}. 
Using the Heidelberg Cuneiform Benchmark Dataset (HeiCuBeDa) \cite{mara2019breaking} a dataset for classification of tablet time period is introduced. 
%
The network created by combining SplineNet \cite{fey2018splinecnn} and PointNet++ \cite{qi2017pointnetplusplus}, and outperforms a PointNet++ network. 
%
%
%

In this paper, we introduce a novel architecture for classifying large-scale point clouds, which outperforms Bogacz et al. \cite{bogacz2020period}. The network is based on DGCNN \cite{dgcnn}, but we introduce several modifications. The main contributions are an AvgPool guided MaxPool layer suited for large-scale point clouds, and a bottom-up neighbor computation in geometric space. 
We also introduce the Maximum Attention visualization method, demonstrating that the network attend to relevant features, as shown in Fig.~\ref{fig:activation}. 
Compared with existing methods \cite{zhang2019explaining, huang2019claim} that show the class prediction of each point, our method only shows the points contributing to the prediction.

\begin{figure*}[ht]
    \begin{center}
        \includegraphics[width=0.95\linewidth]{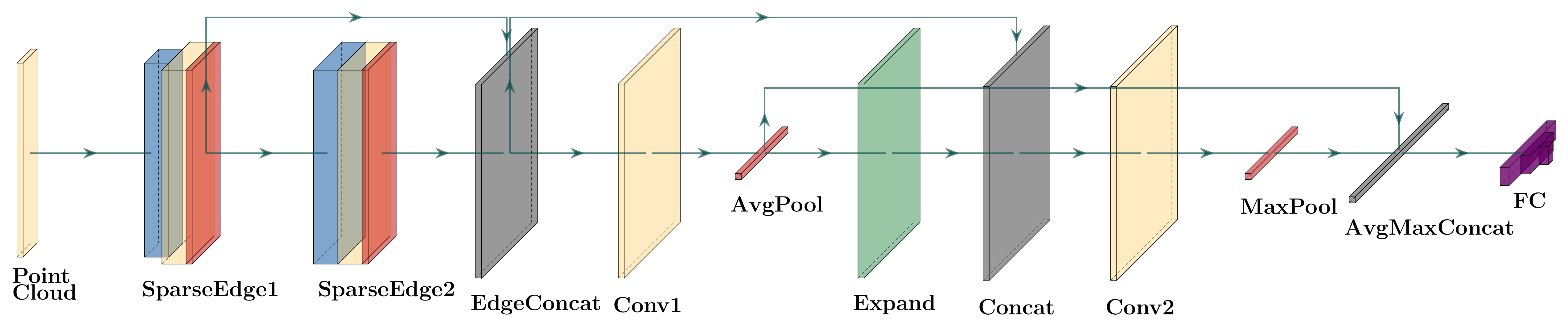}
        \caption{\textbf{Model architecture:} From the input point cloud, the SparseEdge finds neighbors, computes features, and a local MaxPool reduces the output. This process is repeated in the second layer, with neighbors found using the mean neighbor distance in the first layer as the minimum distance.
        The global AvgPool is then computed from the two local MaxPools. The output of the AvgPool is concatenated and combined with the local MaxPools, and a global MaxPool is computed. The AvgPool and MaxPool layers are concatenated, and the fully connected layers compute the class prediction. }
        \label{fig:network}
        \vspace{-5mm}
    \end{center}
\end{figure*}


\section{Method}
\label{sec:method}


The developed network is based on DGCNN \cite{dgcnn}. However, the point clouds are much larger than those generally processed by DGCNN. 
The large number of points are necessary to include the details of the surface imprints. However, a sub-sampling is required both due to memory constraints, the largest tablet has 2,414,753 vertices, and to keep an equal amount of points in each point cloud. This is opposed to  Bogacz et al. \cite{bogacz2020period}  which keep a varying point cloud size. The sub-sampling is set to 32,768 as it is the largest doubling of PointNets original 1024 points, which also encompass the smallest tablet at 44,629 and fit within memory constraints. Four changes have been made to adjust the network to the large number of points.
The network was trained on a Tesla V100-SXM2 with 16 GB memory. 
A standard DGCNN network with 32,768 points is too large to fit onto the GPU. The network was therefore reduced to allow training on the large point cloud. The batch size also had to be reduced to one, to fit the network onto the GPU
%
%

The Edge feature is replaced with the SparseEdge feature \cite{hagelskjaer2021bridging}. Whereas the Edge feature finds the twenty nearest neighbors, the SparseEdge feature finds the sixty nearest neighbors and randomly selects twenty points. This gives a broader range, without increasing memory requirements.

Additionally, as the point clouds are very large, the neighbor space for the Edge feature has been changed. 
In the original DGCNN network, neighbors in the first layer are found in geometric space, and for the second layer, neighbors are found in feature space.
However, as the point clouds are very dense, the possibility of neighbors being repeated from the first layer to the second layer increases. 
To avoid this, we compute the second layer neighbors in geometric space but use a minimum distance for the second layer neighbors.
The minimum distance is found by calculating the mean distance to all neighbors in the first layer. An extended range is ensured by excluding all points within the minimum distance.

A MaxPool layer combined with huge point clouds makes back-propagation difficult, as few points are used for training \cite{bogacz2020period}.
In Bogacz et al., the MaxPool layer is replaced with an AvgPool to compensate for this. However, this excludes the discriminating effect of the MaxPool, which can bring focus to specific points. 
To obtain the strengths of both, we use the AvgPool to aid the MaxPool computation. The AvgPool is computed from the combined SparseEdge layers. It is then expanded and concatenated with the SparseEdge layers, and from this, the MaxPool is computed. Finally, the AvgPool and MaxPool layers are concatenated, and the fully connected layers compute the class predictions. The complete network structure is shown in Fig.~\ref{fig:network}.

The large data size restricts the network to training with a batch size of one. As a result, batch normalization cannot be used. To obtain the advantages of normalization, we instead use group normalization \cite{wu2018group}. Instead of normalizing across batches, the normalization is performed across groups in the channels. This approach has shown good performance when batch normalization is not possible \cite{wu2018group}. 




\subsection{Compensating Class Imbalance}
\label{sec:method:weights}

The class distribution of the dataset introduced in Bogacz et al. is heavily imbalanced \cite{bogacz2020period}. To address this, Bogacz et al. fix each class to 100 samples \cite{bogacz2020period}, excluding much of the data. 

However, several methods exist to compensate for uneven sample distributions \cite{wang2017learning,huang2016learning}.
The general procedure weighs the training loss to provide each class equal attention. 
In this paper, we implement the Inverse Number of Samples strategy \cite{wang2017learning}. This method computes the loss-weight of each sample as a normalized reciprocal of the sample size.

\subsection{Maximum Attention Visualization}
\label{sec:method:saliency}

While test data is used to verify the network's generalization ability, it does not show that the network attend to the intended features. 
To visualize the network's attention, a novel saliency method has been developed. 
Our visualization focuses on the points contributing to the MaxPool, as only these points contribute to the final classification. By separating the MaxPool feature, the forward propagation of the fully connected layer can be used to calculate the saliency. 
The resulting class contribution can be calculated by increasing a value of the MaxPool feature, running a forward-pass, and measure the resulting change in prediction value. 
Furthermore, the major contributors for a class can be determined by setting a cutoff value in relation to the change in prediction. 
Positive contributions are colored green, negative red, and points with a contribution under the cutoff value are colored blue.
We name this visualization method Maximum Attention, as only the points which contribute to the MaxPool are shown. 

\section{Experiments}
\label{sec:experiments}

Several tests are performed to demonstrate the performance of the developed method.
Firstly our method is compared with Bogacz et al. using the same dataset \cite{bogacz2020period}. Secondly, the weight distribution allows us to sample a larger uneven dataset on which our method is trained. Additionally, two new classification datasets have been made concerning tablet metadata. 
These tests can help with classifying tablets when experts are not available. Only UR III data is used for these experiments to simplify the task, as these are the most numerous type.

All experiments are performed using the same architecture and the same hyper-parameters. The hyper-parameters are similar to Bogacz et al. \cite{bogacz2020period}, i.e. 200 epochs, batch size one, and trained with the ADAM optimizer \cite{kingma2014adam}. The only difference is the learning rate which decays from $10^{-3}$ to $10^{-7}$.





\subsection{Performance on the comparison dataset}
\label{sec:experiments:main}


A comparison is performed with the dataset introduced by Bogacz et al. The training and test splits are obtained from the original authors. 
The dataset consists of tablets from four different time periods. 
For each time period, the sample size is 100, except for the Old Assyrian, where 37 are present. 

\begin{figure}[t]
    \begin{center}
        \includegraphics[width=0.85\linewidth]{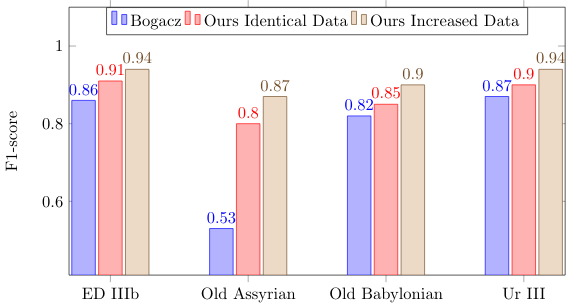}
        \caption{F1-score for each class. Results are shown for Bogacz, ours with 
        identical training data, and increased training data.}
        \vspace{-5mm}
        \label{fig:results_canvas}
    \end{center}
\end{figure}

While the original papers duplicate the Old Assyrian tablets to have 100, the loss-weighing allows us to keep it at 37. The Inverse Number of Samples set the Old Assyrian loss-weight to 1.9 and the others to 0.7. The final F1-score on the test set is 0.88, which outperforms Bogacz et al. at 0.84.
Bogacz et al. also tested a Raster baseline and a reduced PointNet++ \cite{qi2017pointnetplusplus}, that obtains an F1-Score of 0.68 and 0.62 respectively, which our method outperforms.
The F1-scores of our method for each class is shown in Fig.~\ref{fig:results_canvas}, as \textbf{Ours Identical Data}. It is seen that our method outperforms Bogacz et al., shown as \textbf{Bogacz}, on all classes, especially the under-represented class of Old Assyrian, with a difference of 0.53 to 0.80 in F1-score.

As the loss-weighting allows uneven sample sizes, the fixed sample size is unnecessary. To take advantage of this, new training data is sampled using all available tablets, excluding the test data. Additionally, tablets too large, according to Bogacz et al., are also excluded \cite{bogacz2020period}. 
The new dataset now includes 631 tablets, with 312 from UR III. The very large unbalance set the loss-weight of UR III to 0.3, where Old Assyrian is at 2.3. 
The resulting F1-score is further improved to 0.92. The results for each class are shown in Fig.~\ref{fig:results_canvas} as \textbf{Ours Increased Data}. It is seen that the heavy unbalance towards UR III does not lead to over-fitting and the F1-score for all classes increases. Again, the smallest class, Old Assyrian, actually increases the most.

\subsection{Classifying seal presence}
\label{sec:experiments:seal}

A different type of classification task is to determine if seal imprints are present on the tablets. Seals are sometimes found impressed on tablets as a signature \cite[p.~27]{walker1987cuneiform}, either impressed on intentional blank spaces or directly over the text.

The dataset is created by using transliterations from HeiCuBeDa \cite{mara2019breaking} and searching for the "@seal" tag. Tablets with no transliteration are discarded. This results in 136 tablets with seal impressions and 223 without, using ten percent for testing. The test results are shown in Tab.~\ref{tab:seal}.
The performance shows that the network can generalize to this task, but further improvements can be obtained on this dataset.



\begin{table}[ht]
    \centering
        \begin{tabular}{l|l|c|c|}
        \multicolumn{2}{c}{}&\multicolumn{2}{c}{Predicted Label}\\
        \cline{3-4}
        \multicolumn{2}{c|}{} & {w/ seal} & {w/o seal}\\
        \cline{2-4}
        \multirow{2}{*}{True Label}& {w/ seal} & 10 & 2 \\
        \cline{2-4}
        & {w/o seal} & 1 & 22 \\
        \cline{2-4}
        \end{tabular}
    \caption{Seal Presence Classification}
    \label{tab:seal}
    \vspace{-5mm}
\end{table}


\subsection{Classifying left side sign presence}
\label{sec:experiments:left}

\begin{figure}[t]
    \begin{center}
        \includegraphics[trim={8cm 3cm 8cm 3cm},clip,width=0.95\linewidth]{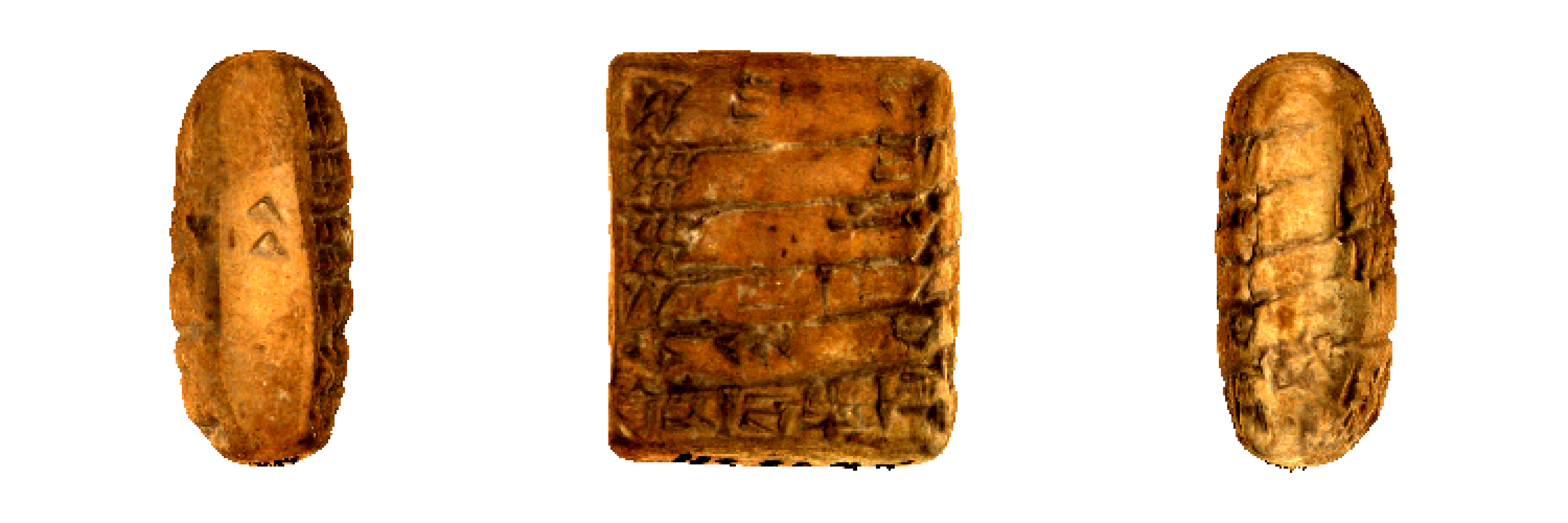} 
        
        \includegraphics[trim={8cm 3cm 8cm 3cm},clip,width=0.95\linewidth]{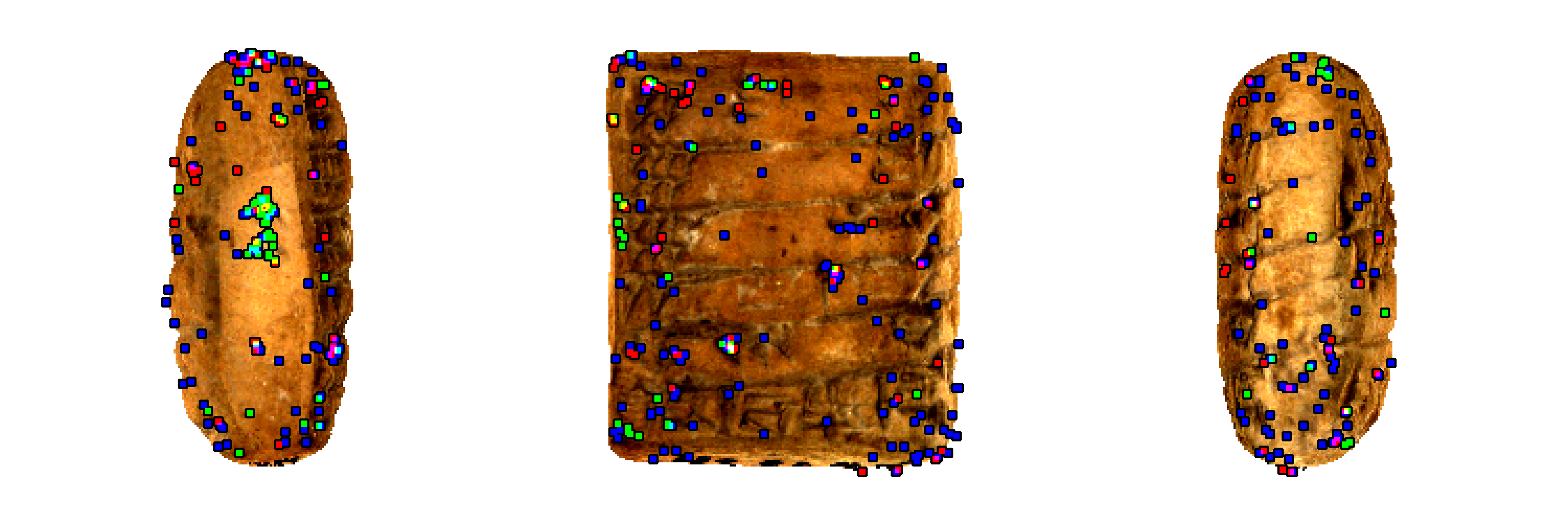} %
        \caption{\textbf{Top:} Tablet "HS 2275" shown from left/middle/right. \textbf{Bottom:} Maximum Attention overlaid according to Sec.~\ref{sec:method:saliency}. Best viewed digitally.}       
        \label{fig:detection2}
        \vspace{-5mm}
    \end{center}
\end{figure}

Another task we introduce is to classify if signs have been written on the left side of the tablet.
For some tablets, the writing is not only on the front and back of the tablet but also on the left side \cite{tsouparopoulou2013reflections}. 
This writing does not follow the general writing direction but is rotated ninety degrees. 
The task is made difficult by the fact that the general writing often overlaps onto the left side \cite[p.~14]{walker1987cuneiform}. The network, therefore, needs to discriminate the sign orientation. An additional challenge is that very few signs can be present, as shown in Fig.~\ref{fig:detection2}.
The dataset is created by using transliterations from HeiCuBeDa \cite{mara2019breaking} and searching for the "@left" tag. The resulting training set is heavily unbalanced as 48 tablets have left side signs, and 276 do not.
The performance on the test set is shown in Tab.~\ref{tab:left}. The network is able to classify most tablets correctly but makes two errors. 

\begin{table}[ht]
    \centering
        \begin{tabular}{l|l|c|c|}
        \multicolumn{2}{c}{}&\multicolumn{2}{c}{Predicted Label}\\
        \cline{3-4}
        \multicolumn{2}{c|}{} & {w/ text} & {w/o text}\\
        \cline{2-4}
        \multirow{2}{*}{True Label}& {w/ text} & 6 & 1 \\
        \cline{2-4}
        & {w/o text} & 1 & 27 \\
        \cline{2-4}
        \end{tabular}
    \caption{Left Side Text Presence}
    \label{tab:left}
\end{table}

\begin{figure}[t]
    \begin{center}
        \includegraphics[trim={8cm 3cm 8cm 3cm},clip,width=0.95\linewidth]{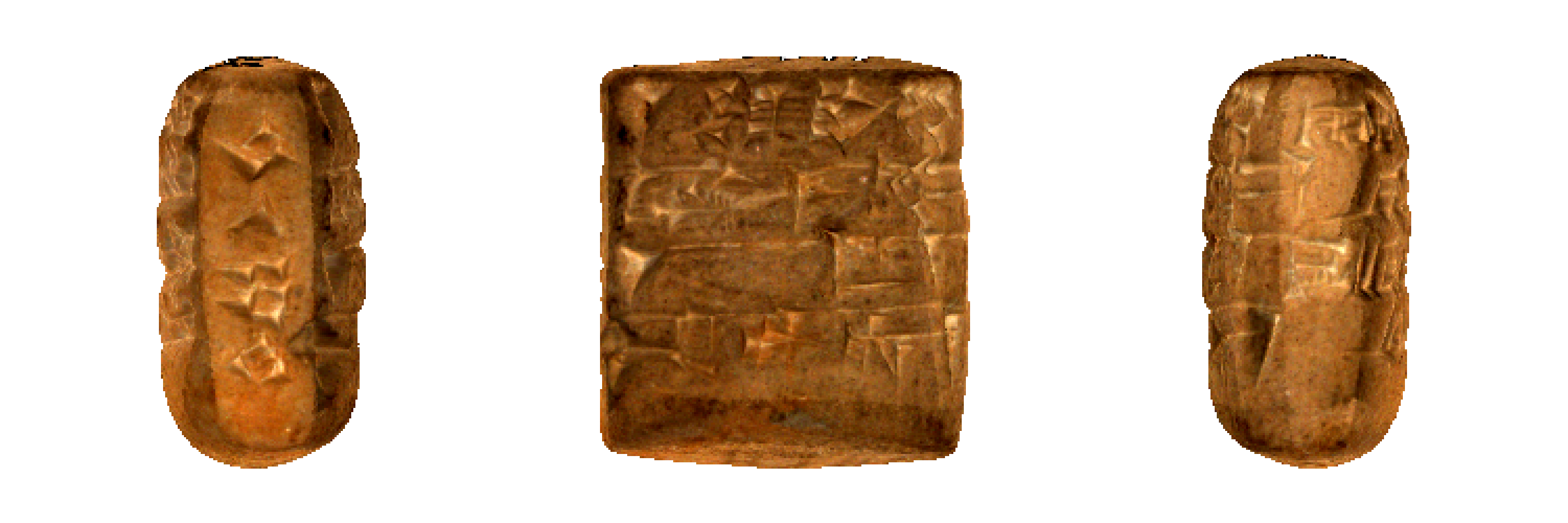}
        
        \includegraphics[trim={8cm 3cm 8cm 3cm},clip,width=0.95\linewidth]{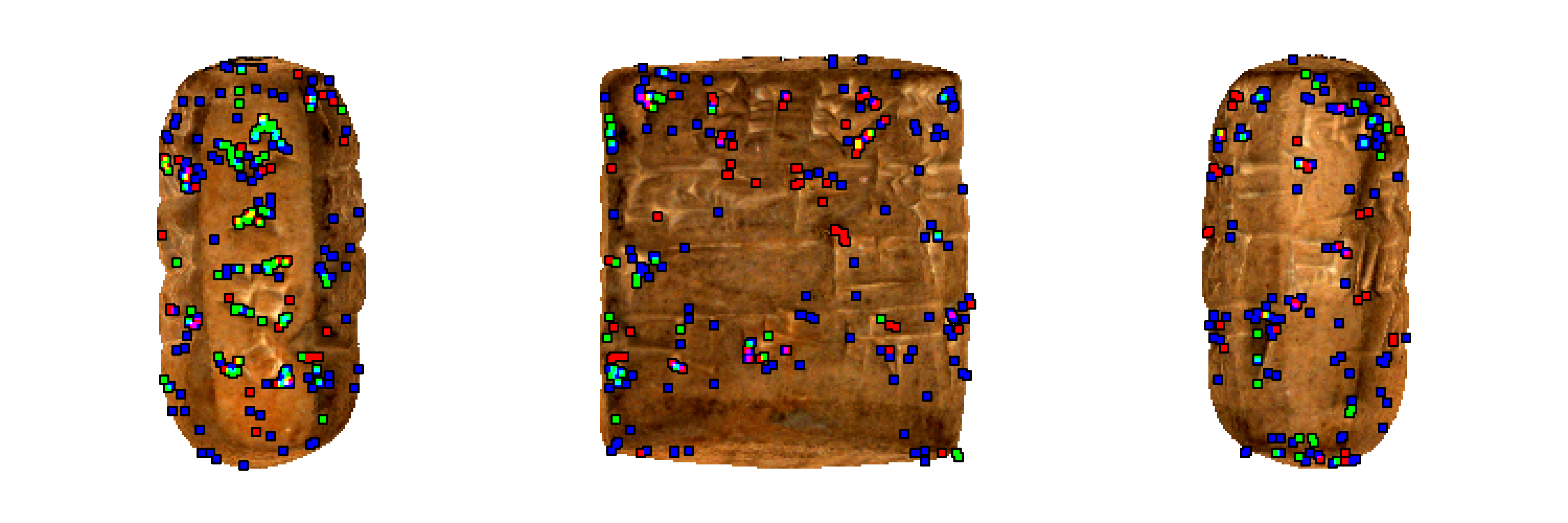}
        \caption{\textbf{Top:} Tablet "HS 2269" shown from left/middle/right. \textbf{Bottom:} Maximum Attention overlaid according to Sec.~\ref{sec:method:saliency}. Best viewed digitally.}
        \label{fig:detection1}
    \end{center}
    \vspace{-5mm}
\end{figure}

To verify that the network focuses on the intended signs the Maximum Activation visualization is used. The activation for two different tablets with left side writing is shown in Fig.~\ref{fig:detection2} and Fig.~\ref{fig:detection1}. It is shown that the network successfully focuses on the actual signs.

        


\subsection{Ablation Study}
\label{sec:ablation}

Several tests are performed to verify the effectiveness of the developed network. All tests are performed on the dataset from Bogacz et al., where our method obtained an average F1-score of 0.88. 

The first test is to verify that the developed network outperforms a classic DGCNN \cite{dgcnn}. As a result of hardware limitations, the DGCNN only has two convolutional layers as opposed to three, and the batch normalization is removed. DGCNN obtains an average F1-score of 0.83, which is far outperformed by our network. However, the reduced DGCNN outperforms the reduced PointNet++ at 0.62 indicating the importance of the local neighbor information.

An ablation study is performed to test the influence of each contribution.
The result of each ablation is shown in Tab.~\ref{tab:ablation}.
In the first test, the standard neighbor search from DGCNN is used as opposed to our minimum distance, shown as \textbf{MD}.
The second test replaces the SparseEdge feature with the classic Edge feature from the DGCNN, shown as \textbf{SE}. 

A test is also performed without using the MaxPool layer. The AvgPool output is simply used directly by the fully connected layer. This reduced network is also tested by replacing the AvgPool layer with an MaxPool layer. The two test are shown as \textbf{Avg} and \textbf{Max}, respectively.
Finally, the network is tested without group normalization, shown as \textbf{GN}.
It is seen that all parts contribute to the performance. The highest contribution comes from using a MaxPool layer, while the AvgPool in combination also improves performance. The SparseEdge also improves the performance significantly.

\begin{table}[h]
    \centering
    \begin{tabular}{|c|c|c|c|c|c|c|}
    \hline
         \textbf{Omitted} & \textbf{None} & \textbf{MD} & \textbf{SE} & \textbf{Max} & \textbf{Avg} & \textbf{GN} \\
    \hline
         \textbf{F1-score} &  0.88 & 0.87 & 0.85 & 0.82 & 0.86 & 0.87 \\
    \hline
    \end{tabular}
    \caption{Resulting F1-scores of ablation study.}
    \label{tab:ablation}
    \vspace{-5mm}
\end{table}

\section{Conclusion and future work}
\label{sec:conclusion}

In this work, we presented a novel network architecture for the classification of large-scale point clouds. 
The developed network is tested on a comparison dataset where it achieves state-of-the-art performance. 
We also introduce two new classification tasks that demonstrate the network's abilities.
Finally, the Maximum Attention visualization method is introduced,
which confirms that the network targets the intended features.
The networks greatest limitation is the memory constraints, restricting the network size. 
The part consuming the largest amount of memory is the adjacency matrix which is used to compute neighbors, as the size is $N^2$, with $N=32,768$ points. 
With access to a GPU with more memory, the performance of a full-size network could be computed. Another interesting aspect is the development of networks able to compute the neighbor information without requiring the large adjacency matrix
The generalizability of the network could also be tested by introducing other tablet databases \cite{cohen2004iclay,anderson2002unwrapping}.




\clearpage

\bibliographystyle{IEEEbib}
\bibliography{refs}

\begin{thebibliography}{10}

\bibitem{walker1987cuneiform}
Christopher Bromhead~Fleming Walker,
\newblock {\em Cuneiform}, vol.~3,
\newblock Univ of California Press, 1987.

\bibitem{cohen2004iclay}
Jonathan~D Cohen, Donald~D Duncan, Dean Snyder, Jerrold Cooper, Subodh Kumar,
  Daniel~V Hahn, Yuan Chen, Budirijanto Purnomo, and John Graettinger,
\newblock ``iclay: Digitizing cuneiform.,''
\newblock in {\em VAST}, 2004, pp. 135--143.

\bibitem{zampieri2019report}
Marcos Zampieri, Shervin Malmasi, Yves Scherrer, Tanja Samard{\v{z}}i{\'c},
  Francis Tyers, Miikka Silfverberg, Natalia Klyueva, Tung-Le Pan, Chu-Ren
  Huang, Radu~Tudor Ionescu, et~al.,
\newblock ``A report on the third vardial evaluation campaign,''
\newblock Association for Computational Linguistics, 2019.

\bibitem{gordin2020reading}
Shai Gordin, Gai Gutherz, Ariel Elazary, Avital Romach, Enrique Jim{\'e}nez,
  Jonathan Berant, and Yoram Cohen,
\newblock ``Reading akkadian cuneiform using natural language processing,''
\newblock {\em PloS one}, vol. 15, no. 10, pp. e0240511, 2020.

\bibitem{bogacz2020period}
Bartosz Bogacz and Hubert Mara,
\newblock ``Period classification of 3d cuneiform tablets with geometric neural
  networks,''
\newblock in {\em 2020 17th International Conference on Frontiers in
  Handwriting Recognition (ICFHR)}. IEEE, 2020, pp. 246--251.

\bibitem{mara2019breaking}
Hubert Mara and Bartosz Bogacz,
\newblock ``Breaking the code on broken tablets: The learning challenge for
  annotated cuneiform script in normalized 2d and 3d datasets,''
\newblock in {\em 2019 International Conference on Document Analysis and
  Recognition (ICDAR)}. IEEE, 2019, pp. 148--153.

\bibitem{anderson2002unwrapping}
Sean~E. Anderson and Marc Levoy,
\newblock ``Unwrapping and visualizing cuneiform tablets,''
\newblock {\em IEEE Computer Graphics and Applications}, vol. 22, no. 6, pp.
  82--88, 2002.

\bibitem{qi2017pointnet}
Charles~R Qi, Hao Su, Kaichun Mo, and Leonidas~J Guibas,
\newblock ``Pointnet: Deep learning on point sets for 3d classification and
  segmentation,''
\newblock in {\em Proceedings of the IEEE conference on computer vision and
  pattern recognition}, 2017, pp. 652--660.

\bibitem{guo2020deep}
Yulan Guo, Hanyun Wang, Qingyong Hu, Hao Liu, Li~Liu, and Mohammed Bennamoun,
\newblock ``Deep learning for 3d point clouds: A survey,''
\newblock {\em IEEE transactions on pattern analysis and machine intelligence},
  2020.

\bibitem{bello2020deep}
Saifullahi~Aminu Bello, Shangshu Yu, Cheng Wang, Jibril~Muhmmad Adam, and
  Jonathan Li,
\newblock ``Deep learning on 3d point clouds,''
\newblock {\em Remote Sensing}, vol. 12, no. 11, pp. 1729, 2020.

\bibitem{dgcnn}
Yue Wang, Yongbin Sun, Ziwei Liu, Sanjay~E. Sarma, Michael~M. Bronstein, and
  Justin~M. Solomon,
\newblock ``Dynamic graph cnn for learning on point clouds,''
\newblock {\em ACM Transactions on Graphics}, 2019.

\bibitem{qi2017pointnetplusplus}
Charles~Ruizhongtai Qi, Li~Yi, Hao Su, and Leonidas~J Guibas,
\newblock ``Pointnet++: Deep hierarchical feature learning on point sets in a
  metric space,''
\newblock in {\em Advances in neural information processing systems}, 2017, pp.
  5099--5108.

\bibitem{fey2018splinecnn}
Matthias Fey, Jan~Eric Lenssen, Frank Weichert, and Heinrich M{\"u}ller,
\newblock ``Splinecnn: Fast geometric deep learning with continuous b-spline
  kernels,''
\newblock in {\em Proceedings of the IEEE Conference on Computer Vision and
  Pattern Recognition}, 2018, pp. 869--877.

\bibitem{zhang2019explaining}
Binbin Zhang, Shikun Huang, Wen Shen, and Zhihua Wei,
\newblock ``Explaining the pointnet: What has been learned inside the
  pointnet?,''
\newblock in {\em CVPR Workshops}, 2019, pp. 71--74.

\bibitem{huang2019claim}
Shikun Huang, Binbin Zhang, Wen Shen, and Zhihua Wei,
\newblock ``A claim approach to understanding the pointnet,''
\newblock in {\em Proceedings of the 2019 2nd International Conference on
  Algorithms, Computing and Artificial Intelligence}, 2019, pp. 97--103.

\bibitem{hagelskjaer2021bridging}
Frederik Hagelskj{\ae}r and Anders~Glent Buch,
\newblock ``Bridging the reality gap for pose estimation networks using
  sensor-based domain randomization,''
\newblock in {\em Proceedings of the IEEE/CVF International Conference on
  Computer Vision}, 2021, pp. 935--944.

\bibitem{wu2018group}
Yuxin Wu and Kaiming He,
\newblock ``Group normalization,''
\newblock in {\em Proceedings of the European conference on computer vision
  (ECCV)}, 2018, pp. 3--19.

\bibitem{wang2017learning}
Yu-Xiong Wang, Deva Ramanan, and Martial Hebert,
\newblock ``Learning to model the tail,''
\newblock in {\em Proceedings of the 31st International Conference on Neural
  Information Processing Systems}, 2017, pp. 7032--7042.

\bibitem{huang2016learning}
Chen Huang, Yining Li, Chen~Change Loy, and Xiaoou Tang,
\newblock ``Learning deep representation for imbalanced classification,''
\newblock in {\em Proceedings of the IEEE conference on computer vision and
  pattern recognition}, 2016, pp. 5375--5384.

\bibitem{kingma2014adam}
Diederik~P Kingma and Jimmy Ba,
\newblock ``Adam: A method for stochastic optimization,''
\newblock {\em arXiv preprint arXiv:1412.6980}, 2014.

\bibitem{tsouparopoulou2013reflections}
Christina Tsouparopoulou,
\newblock ``Reflections on paratextual markers and graphic devices in ur iii
  administrative documents,''
\newblock {\em Textual Cultures}, vol. 8, no. 2, pp. 1--14, 2013.

\end{thebibliography}

\end{document}